\documentclass{article}
\usepackage{ijcai25}

\usepackage{times}
\usepackage{soul}
\usepackage{url}
\usepackage[hidelinks]{hyperref}
\usepackage[utf8]{inputenc}
\usepackage[small]{caption}
\usepackage{graphicx}
\usepackage{amsmath}
\usepackage{amsthm}
\usepackage{booktabs}
\usepackage{algorithm}
\usepackage{algorithmic}
\usepackage[switch]{lineno}

\usepackage{amsfonts}       
\usepackage{amssymb}
\usepackage{amsmath}
\usepackage{amsthm}
\usepackage{arydshln}
\usepackage{booktabs}       
\usepackage{caption}
\usepackage{enumitem}
\usepackage{latexsym}
\usepackage{mathtools}
\usepackage{microtype}      
\usepackage{multirow}
\usepackage{nicefrac}       
\usepackage{subfigure}
\usepackage{threeparttable}
\usepackage[table]{xcolor}
\usepackage{caption} 

\title{Rethinking Remaining Useful Life Prediction with Scarce Time Series Data:\\ Regression under Indirect Supervision}

\author{
Jiaxiang Cheng
\and
Yipeng Pang\and
Guoqiang Hu\\
\affiliations
School of Electrical and Electronic Engineering, Nanyang Technological University, Singapore
\emails
jiaxiang002@e.ntu.edu.sg,
yppang@ntu.edu.sg,
gqhu@ntu.edu.sg
}

\begin{document}

\maketitle

\begin{abstract}
Supervised time series prediction relies on directly measured target variables, but real-world use cases such as predicting remaining useful life (RUL) involve indirect supervision, where the target variable is labeled as a function of another dependent variable. Trending temporal regression techniques rely on sequential time series inputs to capture temporal patterns, requiring interpolation when dealing with sparsely and irregularly sampled covariates along the timeline. However, interpolation can introduce significant biases, particularly with highly scarce data. In this paper, we address the RUL prediction problem with data scarcity as time series regression under indirect supervision. We introduce a unified framework called parameterized static regression, which takes single data points as inputs for regression of target values, inherently handling data scarcity without requiring interpolation. The time dependency under indirect supervision is captured via a parametrical rectification (PR) process, approximating a parametric function during inference with historical posteriori estimates, following the same underlying distribution used for labeling during training. Additionally, we propose a novel batch training technique for tasks in indirect supervision to prevent overfitting and enhance efficiency. We evaluate our model on public benchmarks for RUL prediction with simulated data scarcity. Our method demonstrates competitive performance in prediction accuracy when dealing with highly scarce time series data. 
\end{abstract}

\section{Introduction}

In time series forecasting, the objective is to predict a target value or sequence of target values based on historical observations. Leveraging trending deep learning (DL) techniques, models can be trained end-to-end by mapping historical data to observed values. Subsequently, these trained models utilize the most recent observations to forecast future values. However, a unique scenario arises when the target values during model training cannot be directly measured but are instead approximately labeled based on the observation of a dependent variable. This specific task is categorized as \textit{indirect supervision}~\cite{wang2020learnability}. Within this framework, one of the typical tasks is predicting the remaining useful life (RUL) of subjects, often pertaining to industrial systems. Accurate RUL prediction is crucial for maintenance and decision-making across various industries.

Recent research in RUL prediction under indirect supervision has predominantly focused on \textit{temporal regression} techniques, which utilize sequential time series data to predict target values. For RUL prediction, popular models such as recurrent neural networks (RNN)~\cite{heimes2008recurrent}, convolutional neural networks (CNN)~\cite{sateesh2016deep,yang2019remaining,zhang2020time,leal2022learning,li2022enhanced}, 
long short-term memory (LSTM) or gated recurrent unit (GRU)~\cite{zheng2017long,huang2019bidirectional,lu2020autoencoder,chen2020machine,zhang2020time,li2022enhanced,ince2023joint}, and Transformer or attention-based models~\cite{chen2020machine,li2022enhanced,xu2023new,qin2023interpretable} have been widely adopted. However, challenges arise when dealing with scarce time series data collected \textit{sparsely and irregularly} along the timeline. In such cases, data interpolation becomes necessary to facilitate the implementation of temporal regression models. Nevertheless, interpolation may introduce significant bias, especially when dealing with highly scarce time series data with limited samples~\cite{wang2021deep}. 

In contrast to temporal regression, 
we define \textit{static regression} as the regression of target values solely with the current measurements. Applications with classic models such as multilayer perceptron (MLP)~\cite{leal2022learning}, k-nearest neighbors (kNN)~\cite{viale2023least}, and extreme learning machines (ELM)~\cite{mo2022multi}, fall under this category, assuming that target values depend solely on the samples collected at associated time points. While static regression lacks the ability to capture time dependencies like temporal regression, it is inherently less affected by scarcity in time series data and can make predictions at each time point associated with collected samples without requiring any data treatment.
Moreover, under indirect supervision, when target values are labeled with a time-dependent function, the time dependency is inherently recorded in the timestamp. However, this aspect has not been widely studied in the context of static regression.


In this paper, we first formulate the RUL prediction with data scarcity as a problem of time series regression under indirect supervision and classify different types of time series data based on their sampling frequency. While existing research primarily focuses on temporal regression, we introduce a unified framework named \textit{parameterized static regression}. Unlike trending methods that rely on sequential inputs, a \textit{static regression model} is adopted that operates solely with covariates at associated time points for \textit{posterior estimation}. Subsequently, the parameterized static regression model, denoted with a suffix ``-$f$'', utilizes a \textit{parametrical rectification} (PR) process to capture temporal dependency. This process adjusts \textit{posteriori estimates} from static regression model to fit a parametric function during inference, aligning with the underlying distribution used for labeling during indirect supervision training. We investigate both linear and non-linear parametric functions for RUL labeling during the training process, highlighting the significance of selecting the appropriate labeling function under indirect supervision.

To mitigate overfitting during training, we propose a novel batch training technique specifically designed for indirect supervision. This technique involves sampling from one time series sharing an identical labeling function at a time. By appropriately selecting the \textit{sampling size}, our model can be trained effectively and efficiently, significantly reducing computational costs.  Experimental results demonstrate that our method achieves excellent performance in scenarios of scarce time series data and can outperform both existing static and temporal regression methods in certain cases.
The main contributions in this paper can be summarized as follows: 
\begin{enumerate}
\item We formulate and generalize the RUL prediction task with data scarcity as the regression problem under indirect supervision and formulate various evaluation metrics given different forms of prediction outcomes.
\item By using static regression model for posterior estimation, our method can handle scarce time series data, augmented with the PR process for capturing temporal dependency. 
\item A batch training technique that involves sampling from time series with identical labeling functions is proposed to prevent overfitting and reduce computational costs, particularly with large-scale time series data.
\item Experiments demonstrate the effectiveness of proposed parameterized static regression in handling scarce time series data across benchmarks with simulated data scarcity.
\end{enumerate}

\section{Related Work}

\subsubsection{RUL Prediction under Indirect Supervision} 
RUL prediction is a widely explored task under indirect supervision, with both static and temporal regression methods being extensively proposed. Public benchmarks have been established to evaluate RUL prediction performance~\cite{saxena2008damage,saxena2008turbofan,chao2020aircraft}. In addition to above-mentioned work~\cite{sateesh2016deep,zheng2017long,yang2019remaining,huang2019bidirectional,zhang2020time,chen2020machine,zhang2020time,leal2022learning,li2022enhanced,xu2023new,qin2023interpretable}, auto-encoder (AE) has also been utilized in this context, with approaches like AE-GRU~\cite{lu2020autoencoder} and joint autoencoder-regression~\cite{ince2023joint} integrating auto-encoders with GRU and LSTM, respectively. However, these methods still rely on temporal inputs, limiting their generalizability to scarce time series data.
A modified kNN Interpolation (kNNI) integrated with least squares smoothing (LSS)~\cite{viale2023least} was proposed as a static regression method, where LSS performs a similar function to our proposed PR process. However, this approach only considers linear functions, and the limitations of rectification with function approximation were not studied under different scenarios.

\subsubsection{Handling Data Scarcity in RUL Prediction} 
To the best of our knowledge, only one research investigated the scarcity in time series data for RUL prediction, despite its prevalence in practical applications.
Sparse FMLP~\cite{wang2021deep} extends the functional MLP (FMLP)~\cite{rossi2002functional,rossi2005representation} and addresses data scarcity by leveraging functional principle component analysis (FPCA)~\cite{yao2005functional,happ2018multivariate}. 
The study compares regression performance on scarce time series data with LSTM, a temporal regression method thus using various interpolation methods such as cubic spline, Gaussian process regression (GP), and FPCA~\cite{wang2021deep}. However, this study solely simulated and analyzed the impacts of scarce data during training, overlooking its effects during inference, which is more common in practical scenarios where data scarcity occurs during both stages.  Additionally, the impact of uncertainty arising from the randomness in data scarcity on model performance has not been investigated. Therefore, we propose methods to address the highlighted gaps and evaluate performance with comprehensive experimental results.




\begin{figure}[!t]\centering
\subfigure[RSTS]{
\centering
\includegraphics[width=1.6in]{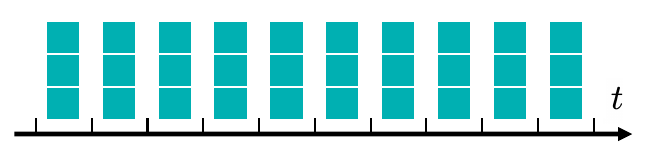}
}%
\hfil
\subfigure[RMTS]{
\centering
\includegraphics[width=1.6in]{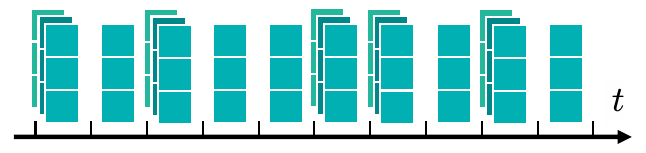}
}%
\hfil
\subfigure[SSTS]{
\centering
\includegraphics[width=1.6in]{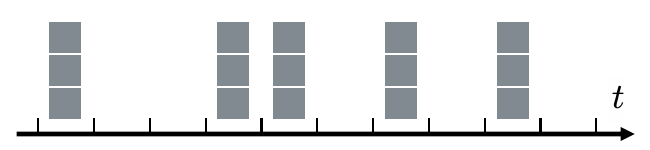}
}%
\hfil
\subfigure[SMTS]{
\centering
\includegraphics[width=1.6in]{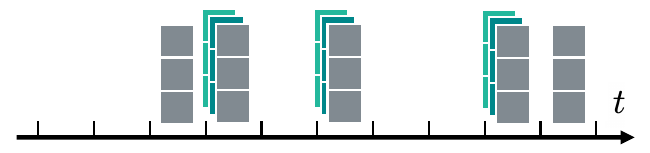}
}%
\caption{\textbf{Illustration of different types of multivariate time series data}. They are categorized and defined in this paper based on their sampling frequency, $i.e.$, scarcity.}
\label{fig:ts}
\vspace{-1em}
\end{figure}

\section{Problem Formulation} 

We begin by formulating various types of time series data, considering the issue of data scarcity, and also outline the framework for time series prediction under indirect supervision. 
Our focus lies on multivariate time series datasets for RUL prediction, which can be segmented into multiple segments, each representing an individual subject. 

Suppose we are provided with a time series data $\mathcal{D} \in \mathbb{R}^{M\times V}$, here $M$ represents the total number of samples and $V$ denotes the total number of variables associated with each sample. The dataset can be partitioned as $\mathcal{D}=[\mathcal{D}_1;\mathcal{D}_2;...;\mathcal{D}_N]$, where {$\mathcal{D}_i= [\mathbf{X}^i_1;\mathbf{X}^i_2;...;\mathbf{X}^i_{T_i}] \in \mathbb{R}^{M_i\times V}$} signifies the $i$-th subject comprising $M_i$ samples. Here, $N$ denotes the total number of subjects, $i.e.$, $i \in \{1,2,...,N\} \subset \mathbb{Z}^+$ and $\sum_{i=1}^N M_i = M$. 
We then introduce the $t \in \mathcal{T}= \{1,2,...,T\} \subset \mathbb{Z}^+$ as the index of consecutive \textit{minimal sampling intervals}. All subjects enter the study from the sampling interval at $t=1$, where $T$ represents the maximum possible index of sampling intervals, {$i.e.$, $T = \max_{i \in \{1,2,...,N\}} T_i$}. Thus, {$\mathbf{X}^i_t=[{\mathbf{x}^i_{t,1}};{\mathbf{x}^i_{t,2}};...;{\mathbf{x}^i_{t,S_t^i}}] \in \mathbb{R}^{S_t^i\times V}$} denotes the samples collected at the $t$-th interval.
Here, $S_t^i\geq0$ indicates the total number of samples collected within the interval, and ${\mathbf{x}^i_{t,s}}\in \mathbb{R}^{1\times V}$ is the $s$-th sample collected within the $t$-th interval, where $s \in \{1,2,...,S_t^i\} \subset \mathbb{Z}^+$. When $S_t^i=0$, it implies that no samples were collected during the $t$-th interval for the $i$-th subject.
Therefore, $\mathcal{T}$ is a fixed equally-spaced time grid aligned with all subjects, regardless of whether samples are collected or not. $T_i$ signifies the \textit{latest sampling interval} reached by the $i$-th subject, $i.e.$, $\forall$ $i$, $T_i\leq T$, and $\sum_{i=1}^N\sum_{t=1}^{T_i} S_t^i = M$.
We then can categorize the provided time series data $\mathcal{D}$ as:
\begin{itemize}
    \item Regular single-sample time series (RSTS) data, if $S_t^i=1$, $\forall$ $i \in \{1,2,...,N\}$ with $t \in \{1,2,...,T_i\}$, $i.e.$, exists one and only one sample at each sampling interval;
    \item Regular multi-sample time series (RMTS) data, if  $S_t^i\geq 1$, $\forall$ $i \in \{1,2,...,N\}$ with $t \in \{1,2,...,T_i\}$, and also $\exists$ $i$, $t$, such that $S_t^i>1$;
    \item Scarce single-sample time series (SSTS) data, if  $S_t^i\leq 1$, $\forall$ $i \in \{1,2,...,N\}$ with $t \in \{1,2,...,T_i\}$, and also $\exists$ $i$, $t$, such that $S_t^i=0$;
    \item Scarce multi-sample time series (SMTS) data, if $\exists$ $i \in \{1,2,...,N\}$ with $t \in \{1,2,...,T_i\}$, $S_t^i>1$, and also $\exists$ $i$, $t$, such that $S_t^i=0$.
\end{itemize}
Different categories are also illustrated in Figure \ref{fig:ts}. Under indirect supervision, target values for prediction may be partially or entirely unavailable. In such cases, target values are approximately labeled based on observations of another dependent variable, such as time $t$. Therefore, the objective is to predict a target value, $y^i_t \sim t$, at the $t$-th interval based on samples collected up to that interval, $i.e.$, $\mathcal{D}_i^{(t)}=[\mathbf{X}^i_1;\mathbf{X}^i_2;...;\mathbf{X}^i_{t}]$.

In the case of RUL prediction, models are trained on subjects that have completed their lifetime to predict the RUL of subjects that have not. During training, when the $i$-th subject reaches the end of its life at $T_i$, $\text{RUL}\coloneq 0$, {$\forall \, \mathbf{x}^i_{T_i,s}$ in $\mathbf{X}^i_{T_i}$}. For $t<T_i$, RUL is often assumed to follow a linear degradation, labeled with $\text{RUL}\coloneq T_i-t$ for each {$\mathbf{x}^i_{t,s}$ in $\mathbf{X}^i_t$, $t<T_i$}.
The popular solutions to this problem typically involve using sequential data, $\mathbf{X}^i_{t\leftarrow t-\omega} = [\mathbf{X}^i_{t-\omega};...;\mathbf{X}^i_{t-1};\mathbf{X}^i_{t}]$, to predict the response variable labeled at time $t$, $i.e.$, $y^i_t$, where $\omega \in \mathbb{Z}^+$ is called \textit{sliding window}. With RSTS data, $\mathbf{X}^i_{t\leftarrow t-\omega}=[\mathbf{x}^i_{t-\omega,1};...;\mathbf{x}^i_{t-1,1};\mathbf{x}^i_{t,1}]$ can be directly used as inputs to the model. However, with RMTS data, preprocessing is required to convert $\mathbf{X}^i_{t}\in \mathbb{R}^{S_t^i\times V}$, where $S_t^i>1$, to {${\mathbf{\tilde x}^i_{t}} \in \mathbb{R}^{1\times V}$}. One common approach is to calculate the average values of each variable across different samples. We refer to the regression method using sequential data, $\mathbf{X}^i_{t\leftarrow t-\omega}$, as inputs as \textit{temporal regression}. Instead, we define the method that only uses $\mathbf{x}^i_{t,s}$ for predicting $y^i_t$ as \textit{static regression}.

When addressing scarce time series, $i.e.$, SSTS or SMTS data, {and $\exists$ $i$, $t$, $S^i_t = 0$}, interpolation becomes necessary for facilitating temporal regression. However, interpolation can introduce significant biases, particularly when handling highly scarce data, where $|\mathcal{T}_{i}| \ll T_i$, $\mathcal{T}_{i}=\{t\,|\,S^{i}_{t}\neq 0\}$. While static regression inherently manage scarce data, it may also encounter challenges if $\mathbf{X}^i_{t}$ is empty when predicting $y^i_t$.


\begin{figure*}[t]
\centering
\includegraphics[width=1.9\columnwidth]{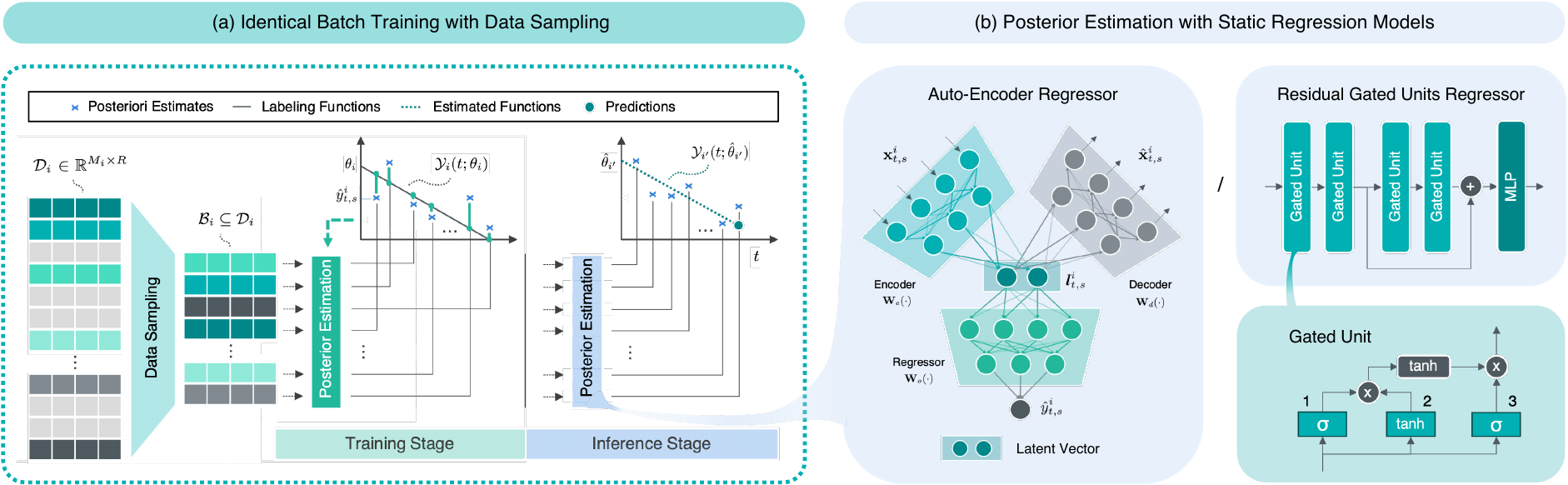} 
\caption{\textbf{Proposed batch training technique and static regression models for posterior estimation}. (a) Batch training technique under indirect supervision conducts data sampling from each $\mathcal{D}_i$ at a time with $|\mathcal{B}_i| =\min(M_i,B)$. (b) Posterior estimation with $\mathbf{x}^i_{t,s}$ using an AER / ResGUR model and $\hat y^{i}_{t,s}$ as outputs (reconstructed $\hat{\mathbf{x}}^i_{t,s}$ also as outputs with AER).}
\label{fig:framework}
\end{figure*}


\section{Methodology}
In this section, we utilize the formulated mathematical problem to introduce the proposed unified framework for handling the scarce time series data under indirect supervision.

\subsection{Parametric Labeling Function}

The selection of the target function for labeling target values is crucial for RUL prediction, as it determines the \textit{underlying distribution} followed by the response variable. 

\subsubsection{Linear Labeling Function}
In the case of RUL prediction, the linear function $y^i_t \coloneq \mathcal{Y}_i(t)=T_i - t$ is widely employed for labeling RUL $w.r.t.$ $t$, where $T_i$ represents the known \textit{end of life} of the $i$-th subject in the training data. Thus, for training purposes, subjects with completed lifetimes are utilized, while a \textit{parametric labeling function} 
is used for labeling as:
\begin{equation}
\mathcal{Y}_i(t;\theta_i)\coloneq \theta_i - t,
\label{eq:linear}
\end{equation}
where $\theta_i$ denotes the known complete lifetime of the $i$-th subject. For prediction, a function $\mathcal{Y}_{i'}(t;\hat\theta_{i'})=\hat\theta_{i'}-t$ is approximated for the $i'$-th subject with an unknown end of life in the test data, where $\hat\theta_{i'}$ corresponds to the estimated lifetime. Additionally, $\mathcal{Y}_{i'}(T_{i'};\hat\theta_{i'})=\hat\theta_{i'}-T_{i'}$ denotes the predicted RUL at the latest lifetime $T_{i'}$ of the $i'$-th subject.

\subsubsection{Non-Linear Labeling Function}
Alternatively, non-linear labeling functions such as the \textit{two-parameter Weibull distribution} are widely employed to model the probability of the \textit{event of interest} occurring by time $t$. The \textit{cumulative distribution function} ($c.d.f.$), denoted as $F(t)$, is expressed as $F(t)=1-\exp({-{(t/\eta)}^\beta})$, where $\beta$ and $\eta$ represent the \textit{shape} and \textit{scale} parameters that specify the distribution. 
When the event of interest is the end of life, the $F(t)$ provides an estimate of the accumulated probability of reaching the end of life, where $1-F(t)$ illustrates the degradation pattern that can approximate the change in RUL. With $\beta$ fixed and adjusting $\eta \propto T_i$ for each $i$-th subject during training, we maintain a one-degree-of-freedom constraint and streamline the function approximation during inference. As $0\leq 1-F(t)\leq 1$, a scaling factor $\alpha$ is introduced to adapt $1-F(t)$ for labeling RUL as $\mathcal{Y}_{i}(t)=\alpha (1-F(t;T_i))$, where $\alpha$ defines the maximum limit of RUL similarly proposed in prior work~\cite{heimes2008recurrent}. Hence, we obtain a non-linear labeling function following the two-parameter Weibull distribution as:
\begin{equation}
\mathcal{Y}_i(t;\theta_i) \coloneq \alpha \exp({-{(t/\theta_i)}^\beta})
\label{eq:weibull}
\end{equation}
where $\theta_i \propto T_i$. An interesting property of the $c.d.f.$ of the two-parameter Weibull distribution is its transformation into a linear function via the equation:
\begin{equation}
\ln(-\ln(1-F(t)))/\beta=\ln t - \ln \eta
\end{equation}
Therefore, approximating $\mathcal{Y}_{i'}(t;\hat\theta_{i'})$ following Equation \ref{eq:weibull} for the $i'$-th subject is equivalent to finding $\tilde \theta_{i'}$ from $\mathcal{Y}_{i'}(\tilde t;\tilde\theta_{i'})= \tilde \theta_{i'} - \tilde t$, where $\mathcal{Y}_{i'}(\tilde t;\tilde\theta_{i'})=-\ln(-\ln(\mathcal{Y}_{i'}(t;\hat\theta_{i'})/\alpha))/\beta$, $\tilde t = \ln t$, and $\tilde \theta_{i'}$ $=\ln \hat\theta_{i'}$. This form mirrors that of the linear function as Equation \ref{eq:linear}.



%

\subsection{Posterior Estimation}

After labeling each sample $\mathbf{x}^i_{t,s}$ of the $i$-th subject sampled at the $t$-th interval 
$y^i_t=\mathcal{Y}_{i}(t;\theta_{i})$, we then train a static regression model to map $\mathbf{x}^i_{t,s} \rightarrow y^i_t$. 

\subsubsection{Identical Batch Training}
As samples from the same $i$-th subject share the identical labeling function $\mathcal{Y}_{i}(t;\theta_{i})$, we propose a novel batch training approach illustrated in Figure \ref{fig:framework}. Each batch input consists of a subset of data: 
\begin{equation}
\mathcal{B}_i \subseteq \mathcal{D}_i, \, |\mathcal{B}_i| =\min(M_i,B),
\label{eq:batch}
\end{equation}
which is sampled from the same $i$-th subject. Here, $B$ represents the \textit{sample size}, and $|\mathcal{B}_i| < B$ when $M_i<B$. 
Sampling is conducted randomly following a uniform distribution, $U(0,1)$.
As multivariate time series data may contain non-informative random noises, incorporating data sampling during the training process can mitigate overfitting to noise and enhance the generalization capability.

\subsubsection{Static Regression Model}
We employ an AER model, illustrated in Figure \ref{fig:framework}. The AE comprises two MLPs serving as \textit{encoder} and \textit{decoder}, with weight matrices denoted as $\mathbf{W}_e(\cdot)$ and $\mathbf{W}_d(\cdot)$, respectively. The encoder maps input features $\mathbf{x}^i_{t,s}$ into a \textit{latent vector} $\textbf{\textit{l}}^i_{t,s}$, while the decoder reconstructs the inputs from the extracted latent features. An additional MLP with weight matrix $\mathbf{W}_o(\cdot)$ generates the output from $\textbf{\textit{l}}^i_{t,s}$. When the decoder is excluded, the AER reduces to a single MLP model, $i.e.$, $\mathbf{W}_o(\mathbf{W}_e(\cdot))$. 
{Leveraging the AE allows for extracting representative information from inputs through dimensionality reduction.}

Additionally, 
we introduce a residual gated units regressor (ResGUR) for posterior estimation, also shown in Figure \ref{fig:framework}.
The computations for this static form of gated unit are as:
\begin{equation}
\mathbf{c}=\sigma(\mathbf{W}_1 \mathbf{x} + \mathbf{b}_1) \odot \text{tanh}(\mathbf{W}_2 \mathbf{x} + \mathbf{b}_2),
\end{equation}
\begin{equation}
\hat{\mathbf{y}}=\sigma(\mathbf{W}_3 \mathbf{x} + \mathbf{b}_3) \odot \text{tanh}(\mathbf{c}),
\end{equation}
where $\odot$ denotes element-wise product, and $\mathbf{W}_n$ and $\mathbf{b}_n$ are the corresponding weight matrices and biases. 


\subsubsection{Loss Function}
During the training of AER, we obtain two different outputs: the estimated $\hat y^i_{t,s}=\mathbf{W}_o(\mathbf{W}_e(\mathbf{x}^i_{t,s}))$ and the reconstructed $\hat{\mathbf{x}}^i_{t,s}=\mathbf{W}_d(\mathbf{W}_e(\mathbf{x}^i_{t,s}))$. The loss obtained from an input batch $\mathcal{B}_i \subseteq \mathcal{D}_i$ is computed with both $\hat y^i_{t,s}$ and $\hat{\mathbf{x}}^i_{t,s}$ using the following loss function:
\begin{equation}
\mathcal{L}=\frac{1}{|\mathcal{B}_i|}\sum_{\mathbf{x}^i_{t,s}\in\mathcal{B}_i}(\hat y^i_{t,s}-\mathcal{Y}_{i}(t;\theta_{i}))^2 + \gamma (\hat{\mathbf{x}}^i_{t,s}-\mathbf{x}^i_{t,s})^2,
\label{eq:loss}
\end{equation}
{where the first term represents the \textit{regression loss}, ensuring regression accuracy, while the second term is the \textit{reconstruction loss}, which ensures the effectiveness of the latent space representation.
The parameter $\gamma$ denotes the weight assigned to the reconstruction loss. When $\gamma=0$, the model training is equivalent to training the MLP model, $i.e.$, $\mathbf{W}_o(\mathbf{W}_e(\cdot))$.}
The ResGUR model is trained with only the \textit{regression loss}.

\begin{table*}[!t]
\caption[Statistics of 13 Subsets in CMAPSS (RSTS) and N-CMAPSS (RMTS)]{\textbf{Statistics of 13 subsets in CMAPSS (\textit{RSTS}) and N-CMAPSS (\textit{RMTS}).} Each subset contains built-in training and testing data. $\bar{T}_i$ (or $\bar{T}_{i'}$) and $\bar{S}^i_{t}$ (or $\bar{S}^{i'}_{t}$) represent the average values across all subjects.}
\centering\fontsize{9}{11}\selectfont
\begin{tabular}{>{\centering}p{0.13cm}>{\centering}p{0.13cm}>{\centering}p{0.9cm}>{\centering}p{0.9cm}>{\centering}p{0.9cm}>{\centering}p{0.9cm}>{\centering}p{0.8cm}>{\centering}p{0.8cm}>{\centering}p{0.8cm}>{\centering}p{0.8cm}>{\centering}p{0.8cm}>{\centering}p{0.8cm}>{\centering}p{0.8cm}>{\centering}p{0.8cm}>{\centering\arraybackslash}p{0.8cm}} 
\toprule
\multicolumn{2}{c}{\multirow{2}{*}{Data}} &  \multicolumn{4}{c}{CMAPSS}  & \multicolumn{9}{c}{N-CMAPSS} \\
\cmidrule(lr){3-6} \cmidrule(lr){7-15}
\multicolumn{2}{l}{} & FD001 & FD002 & FD003 & FD004 & DS01 & DS02 &DS03&DS04&DS05&DS06&DS07&DS08a&DS08c \\
\midrule
\multirow{4}{*}{\rotatebox{90}{Train}} & $N$ & 100 & 260 & 100 & 249 & 6 & 6 & 9 & 6 & 6 & 6 & 6 & 9 & 6 \\
   & $M$ & 20.6K & 53.8K & 24.7K & 61.2K & 4.9M & 5.3M & 5.6M & 6.4M & 4.4M & 4.3M & 4.4M & 4.9M & 4.3M \\
    & $\bar{T}_i$ & 206 & 206 & 247 & 245 & 92 & 74 & 73 & 85 & 81 & 79 & 78 & 67 & 52 \\
  & $\bar{S}^i_{t}$ & 1 & 1 & 1 & 1 & 8.9K & 11.8K & 8.4K & 12.5K & 8.9K & 9.0K & 9.3K & 8.0K & 13.6K \\
\cmidrule(lr){1-15} 
\multirow{4}{*}{\rotatebox{90}{Test}} & $N'$ & 100 & 259 & 100 & 248 & 4 & 3 & 6 & 4 & 4 & 4 & 4 & 6 & 4 \\
 & $M'$ & 13.1K & 34.0K & 16.6K & 41.2K & 2.7M & 1.3M & 4.3M & 3.6M & 2.6M & 2.5M & 2.9M & 3.7M & 2.1M \\
    & $\bar{T}_{i'}$ & 130 & 131 & 165 & 166 & 85 & 67 & 73 & 86 & 81 & 80 & 86 & 63 & 59 \\
  & $\bar{S}^{i'}_{t}$ & 1 & 1 & 1 & 1 & 8.0K & 6.2K & 9.7K & 10.5K & 7.8K & 7.8K & 8.3K & 9.7K & 8.9K \\
\bottomrule
\end{tabular}
\label{tab:data}
\end{table*}

\subsection{Parametrical Rectification}

During the inference stage, we can obtain $\hat y^{i'}_{t,s}$ from each $\mathbf{x}^{i'}_{t,s}$ of the $i'$-th subject in the testing data from the posterior estimation. Since the labeling function $\mathcal{Y}_{i}(t;\theta_i)$ is employed to determine the target values $y^i_{t}$
during training, we can similarly approximate a function $\mathcal{Y}_{i'}(t;\hat{\theta}_{i'})$ by estimating $\hat{\theta}_{i'}$.

Suppose the maximum lifetime the $i'$-th subject has reached is $T_{i'}$. Then, we have {$\mathcal{D}_{i'} = [\mathbf{X}^{i'}_1;\mathbf{X}^{i'}_2;...;\mathbf{X}^{i'}_{T_{i'}}]\in \mathbb{R}^{M_{i'}\times V} $}, with a total of $M_{i'}$ samples collected up to the $T_{i'}$-th interval. We can simplify and reindex the matrix as:
\begin{equation}
\mathcal{D}_{i'} =[\mathbf{x}^{i'}_{1}, \mathbf{x}^{i'}_{2}, ..., \mathbf{x}^{i'}_{M_{i'}}] \Leftrightarrow   [\mathbf{X}^{i'}_1;\mathbf{X}^{i'}_2;...;\mathbf{X}^{i'}_{T_{i'}}] \,,
\end{equation}
where for every $\mathbf{x}^{i'}_{j}$, there exists $t\in\{1,2,...,T_{i'}\}$ such that {$\mathbf{x}^{i'}_{j}$ in $\mathbf{X}^{i'}_{t}$}. 
From the posterior estimation, we obtain $\hat{\mathbf{y}}_{i'}=[\hat y^{i'}_{1},\hat y^{i'}_{2},...,\hat y^{i'}_{M_{i'}}]$  and the associated indices of sampling intervals $\mathbf{t}_{i'}=[t^{i'}_{1},t^{i'}_{2},...,t^{i'}_{M_{i'}}]$.

\subsubsection{Objective Function}
The objective is to find the $\hat{\theta}_{i'}^*$ that minimizes the sum of squared errors between the estimated target values $\hat y^{i'}_{j}$ and the values of the chosen parametric function $\mathcal{Y}_{i'}(t;\hat{\theta}_{i'})$ at the observed time points $t^{i'}_{j}$ as:
\begin{equation}
\mathcal{J}(\hat{\theta}_{i'} \,|\, \hat{\mathbf{y}}_{i'}, \mathbf{t}_{i'}) = \frac{1}{M_{i'}} \sum_{j=1}^{M_{i'}} (\hat y^{i'}_{j} - \mathcal{Y}_{i'}(t^{i'}_{j};\hat{\theta}_{i'}))^2 .
\end{equation}
The Levenberg-Marquardt algorithm~\cite{gavin2019levenberg} is used for optimization.
Once we obtain the estimated parameter $\hat{\theta}_{i'}$ for the $i'$-th subject using $\hat{\mathbf{y}}_{i'}$ and $\mathbf{t}_{i'}$, we can rectify the predicted target values with the function $\mathcal{Y}_{i'}(t;\hat{\theta}_{i'})$. For example, the predicted RUL at the last sampling interval $T_{i'}$ is rectified as $\mathcal{Y}_{i'}(T_{i'};\hat{\theta}_{i'})$. If $t^{i'}_{M_{i'}}=T_{i'}$, 
$\hat y^{i'}_{M_{i'}}$ is then commonly used as the prediction of $y^{i'}_{T{i'}}$. However, when $t^{i'}_{M_{i'}}\neq T_{i'}$, 
the prediction of $y^{i'}_{T_{i'}}$ cannot be obtained directly without data imputation. In such cases, we use $\mathcal{Y}_{i'}(T_{i'};\hat{\theta}_{i'})$ as the prediction, demonstrating an apparent advantage.

\section{Experiments}


In this section, we conduct extensive experiments to address the following key research questions:

\textbf{Q1} How does our method's overall performance compare to previous approaches for \textbf{\textit{RUL prediction with scarce time series data}} on public benchmarks?

\textbf{Q2} How effectively does the proposed PR process enhance the prediction performance of static regression models?

\textbf{Q3} What is the impact of selecting different labeling functions, \textit{e.g.}, linear vs. non-linear labeling function, on the prediction performance of our method?

\textbf{Q4} How do various configurations, \textit{e.g.}, the sampling size for identical batch training and the reconstruction loss in AER, influence the accuracy of RUL prediction?

\subsection{Dataset Description}

We use two turbofan engine degradation simulation datasets: CMAPSS~\cite{saxena2008turbofan} 
and N-CMAPSS~\cite{chao2020aircraft}, which are commonly employed benchmarks for RUL prediction. 
The subjects in the datasets are the individual \textit{engine units} and each \textit{flight cycle} the units committed is considered as the \textit{minimal sampling interval}.
As summarized in Table \ref{tab:data}, there are built-in train and test data in all subsets.

\textbf{CMAPSS}
In CMAPSS, a typical RSTS data, the units in the train set have completed their lifetime at $T_i$, and the units in the test set are in the middle of their lifetime, with an objective RUL to predict at $T_{i'}$ for each $i'$-th unit, with true $y^{i'}_{T_{i'}}$ provided in the test data. 

\textbf{N-CMAPSS}
In N-CMAPSS, a typical RMTS data, the units in both sets have completed their lifetime, and the objective is to predict the RUL corresponding to each $t\leq T_{i'}$ of each ${i'}$-th unit in the test data, with true $y^{i'}_{t}$ provided. 

For N-CMAPSS, we evaluate the accuracy of the model predictions in different ways given different experiments. 
If we evaluate the \textit{sample-wise} prediction performance, we evaluate accuracy by comparing each $\hat{\mathbf{y}}^{i'}_{t,s}$ projected using data collected up to the $s$-th sample at $t$ with the true label ${\mathbf{y}}^{i'}_{t}$. We can also compare the \textit{interval-wise} prediction accuracy, which requires prediction results at each $t$, denoted as $\hat{\mathbf{y}}^{i'}_{t}$. 
Two different ways are used for performing interval-wise prediction: (1) Take the median of all sample-wise predictions $\hat{\mathbf{y}}^{i'}_{t,s}$ using RMTS or SMTS at the $t$-th interval as $\hat{\mathbf{y}}^{i'}_{t}$; or (2) Convert the inputs as $\tilde{\mathbf{x}}^{i'}_{t}$ with the average values of ${\mathbf{x}}^{i'}_{t,s}$ at $t$, $i.e.$, RSTS or SSTS, thus directly leading to 
$\hat{\mathbf{y}}^{i'}_{t}$. 


Each $v$-th variable, $x_v$, is normalized with z-score normalization as $(x_v-\mu_v)/\sigma_v$, where $\mu_r$ and $\sigma_v$ are the mean and standard deviation of each $v$-th variable.

\begin{table*}[!t]
\caption[Prediction performance (RMSE$_{\,i}$) on CMAPSS under simulated data scarcity \textit{in the train set only}.]{\textbf{Prediction performance (RMSE$_{\,i}$) on CMAPSS under simulated data scarcity \textit{in the train set only}.} 
While Sparse MFMLP shows better performance in nearly half of the experiments, it is only compared between the \textbf{average} performance of our method and the results claimed in~\cite{wang2021deep}, which lacks in uncertainty quantification. Meanwhile, our proposed method shows increasingly significant outperformance \textbf{as scarcity increases}. Our method also demonstrates superior performance with both train and test sets as SSTS, as in Figure~\ref{fig:cmpscarce}.}
\centering\fontsize{9}{11}\selectfont
\begin{tabular}{>{\centering}p{1.8cm}>{\centering}p{2.8cm}>{\centering}p{2.4cm}>{\centering}p{2.4cm}>{\centering}p{2.4cm}>{\centering\arraybackslash}p{2.4cm}} 
\toprule
Scarcity & Method & FD001 &FD002 &  FD003 & FD004 \\
\midrule
 \multirow{4}*{50\%} &  LSTM + GP & $26.45$ &$24.76$ &  $26.43$ &  $25.26$ \\
   & LSTM + MPACE & $17.13$ &$18.52$ &  $19.49$ & $20.14$ \\
  & Sparse MFMLP & $\textbf{15.02}$ & $16.98$ &  $\textbf{14.41}$ &$\textbf{17.19}$ \\
    \cdashline{2-6}
  & \textbf{Ours} & $16.74\,\pm\,0.91$ &$\textbf{16.59}\,\pm\,1.20$ &  $18.91\,\pm\,1.35$ &  $18.57\,\pm\,0.71$ \\
   \hline
 \multirow{4}*{70\%}  &  LSTM + GP & $29.98$ &$30.28$ &  $30.42$  &  $28.91$ \\
  &  LSTM + MPACE & $17.48$ &$18.36$ &  $20.09$  & $21.56$ \\
 &  Sparse MFMLP& $16.85$ &$17.08$ &  $\textbf{16.06}$ & $\textbf{18.18}$\\
    \cdashline{2-6}
 \cellcolor[gray]{1}& \textbf{Ours} & $\textbf{16.82}\,\pm\,0.91$ & $\textbf{16.04}\,\pm\,1.48$ & $18.80\,\pm\,1.29$ & $18.79\,\pm\,0.54$ \\
   \hline
 \multirow{4}*{90\%}  &  LSTM + GP & $34.22$ &$36.78$ &  $35.89$  & $34.33$ \\
   & LSTM + MPACE & $21.85$ &$23.10$ &  $21.10$  & $23.54$ \\
  &  Sparse MFMLP& $20.27$ & $18.67$ & \textbf{19.63}  & $\textbf{19.03}$ \\
  \cdashline{2-6}
 & \textbf{Ours} & $\textbf{19.58}\,\pm\,5.82$ & $\textbf{15.89}\,\pm\,1.05$ & $\textbf{20.45}\,\pm\,\textbf{4.74}$ & $\textbf{19.92}\,\pm\,\textbf{0.89}$ \\
\bottomrule
\end{tabular}
\label{tab:cmscarce}
\end{table*}

\begin{figure*}[!t]\centering
\subfigure[FD001 (ResGUR-$f$)]{
\centering
\includegraphics[width=1.5in]{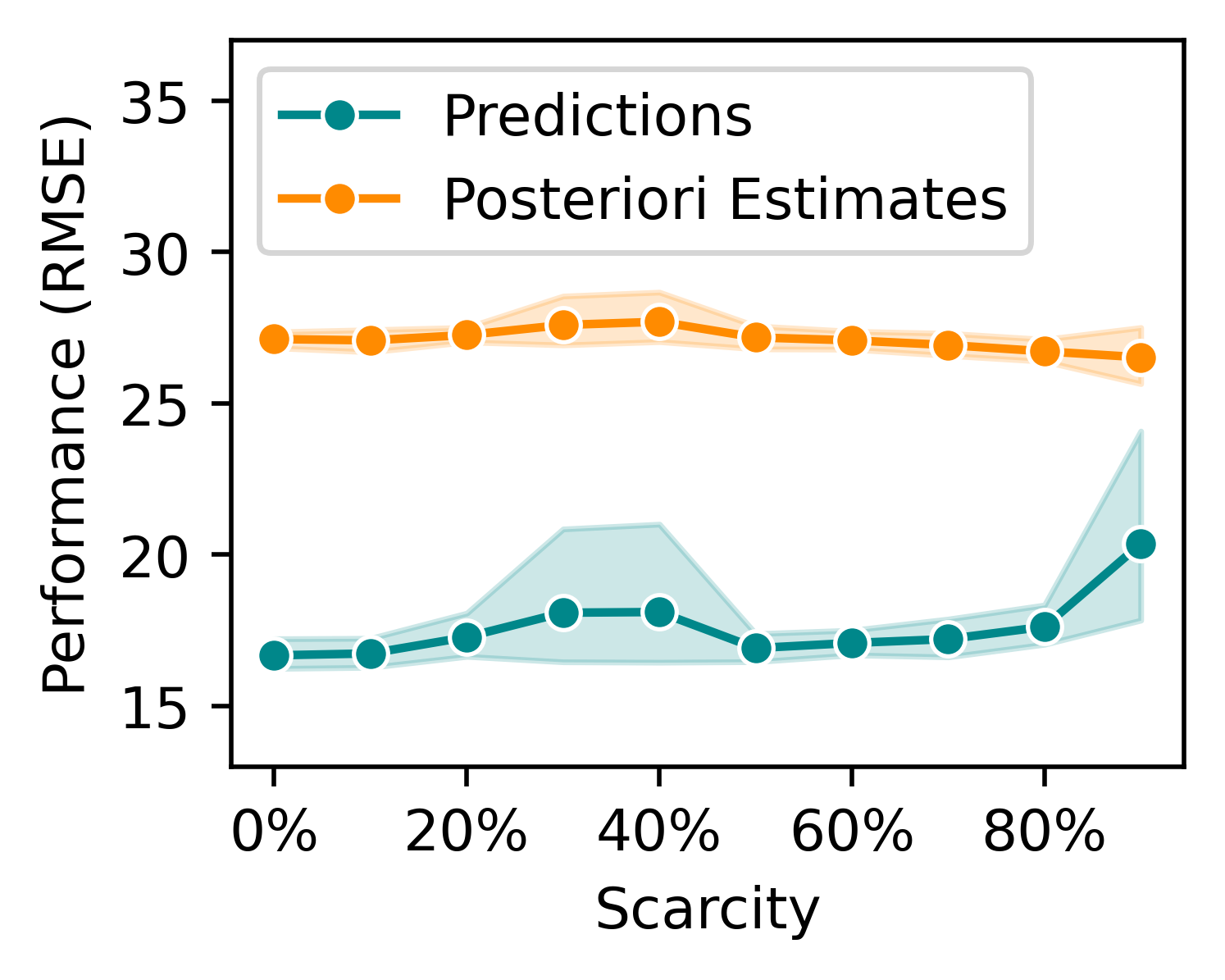}
\label{fig:cmpscarce1}
}%
\hfil
\subfigure[FD003 (ResGUR-$f$)]{
\centering
\includegraphics[width=1.5in]{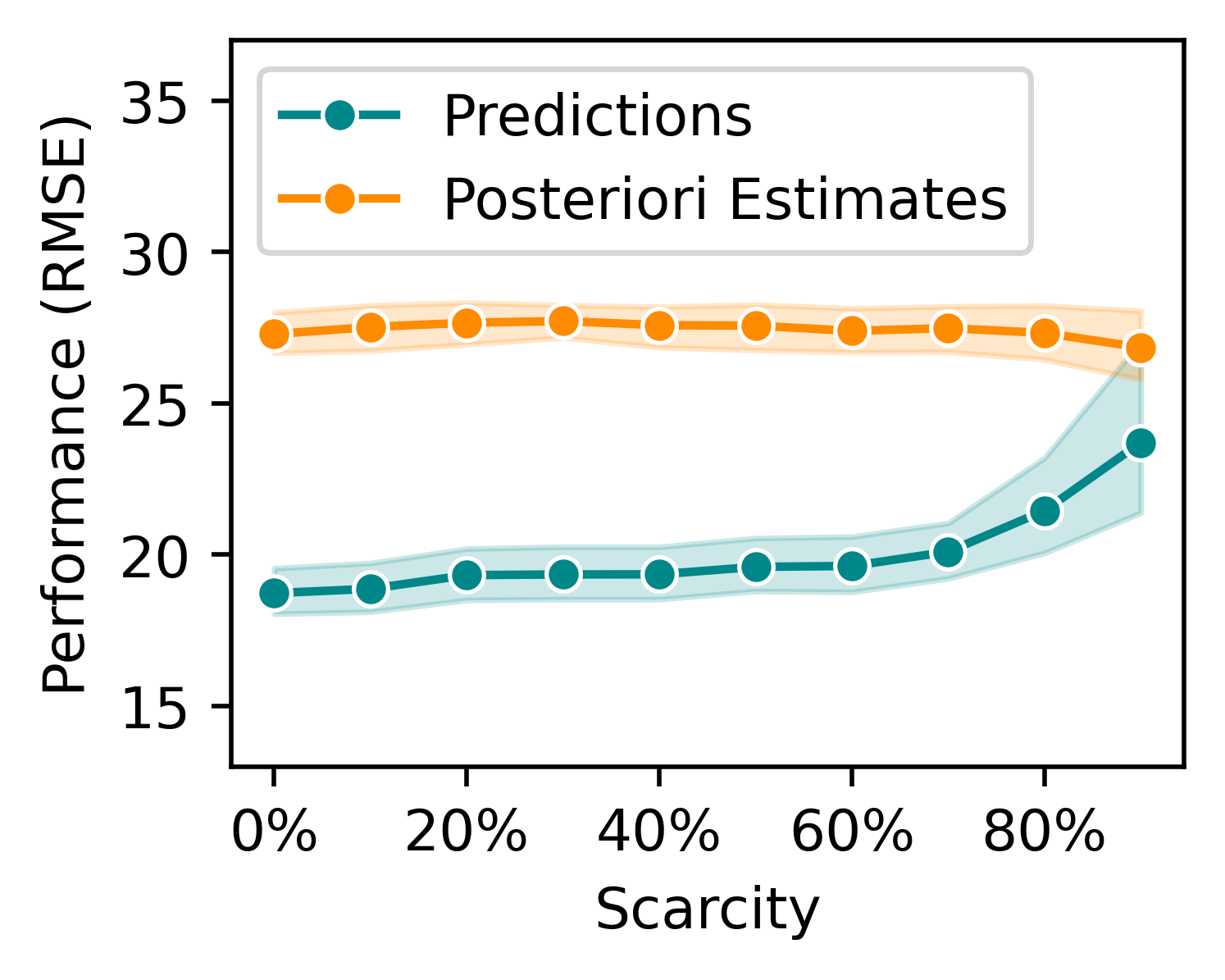}
\label{fig:cmpscarce3}
}%
\hfil
\subfigure[FD002 (AER-$f$)]{
\centering
\includegraphics[width=1.5in]{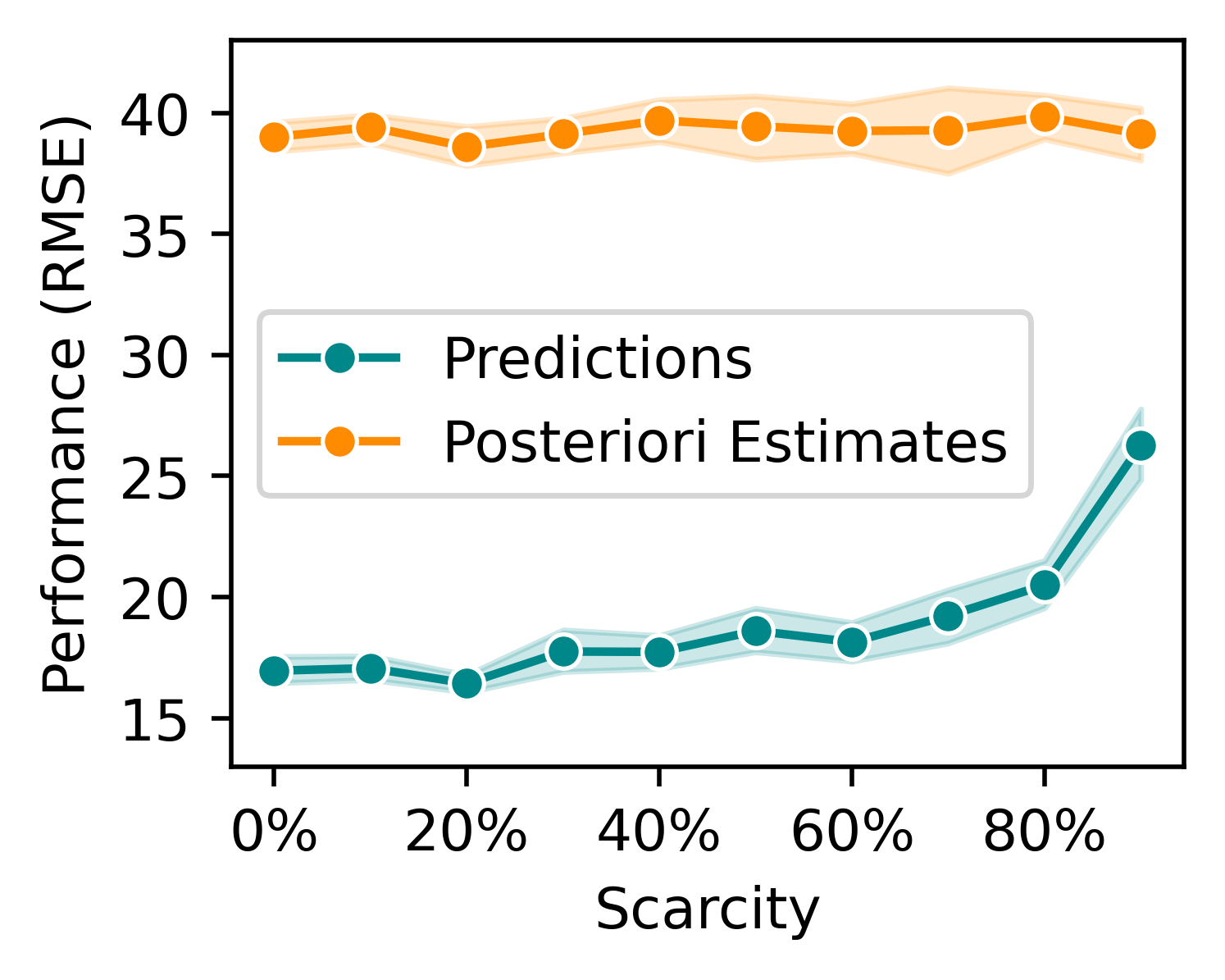}
\label{fig:cmpscarce2}
}%
\hfil
\subfigure[FD004 (AER-$f$)]{
\centering
\includegraphics[width=1.5in]{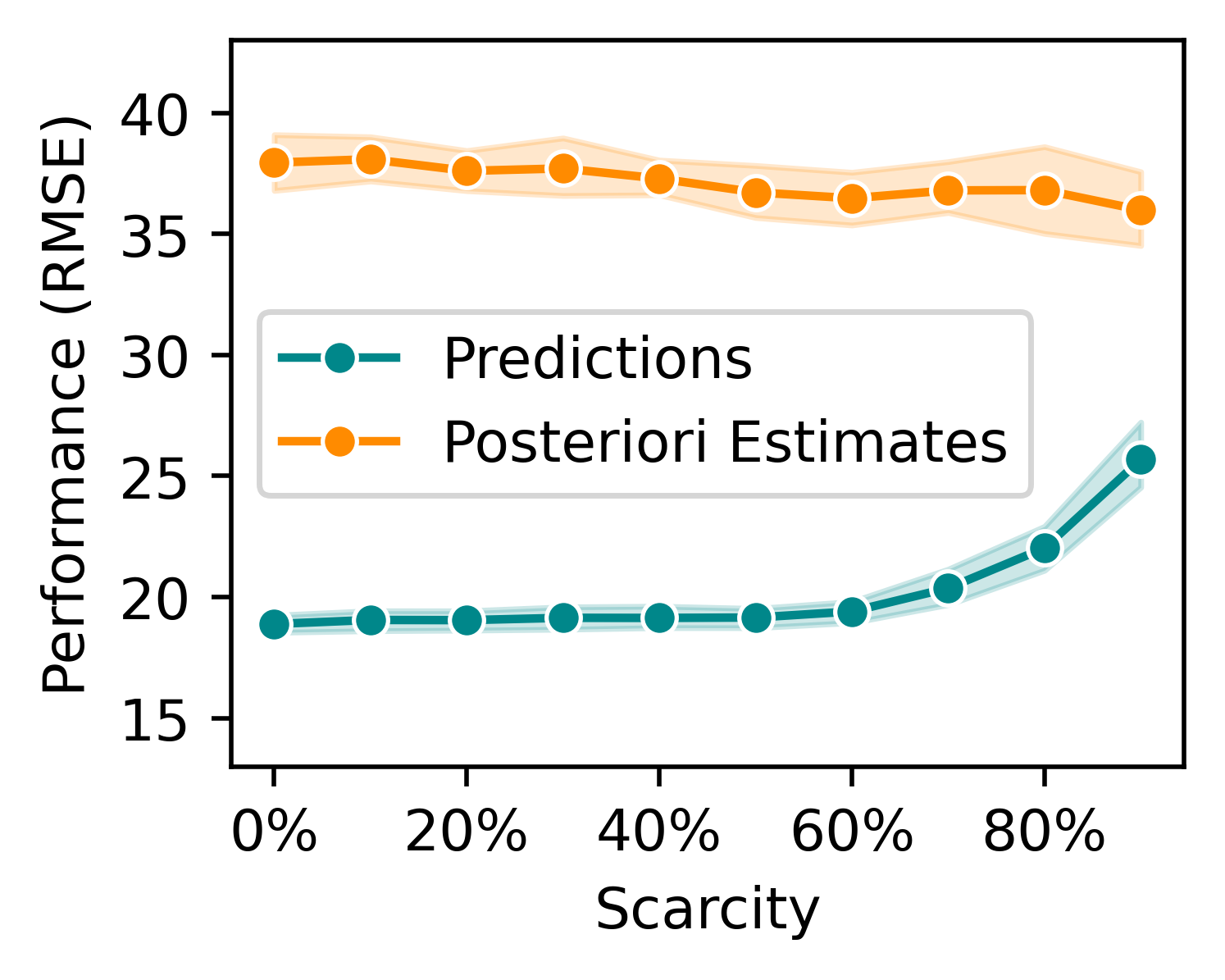}
\label{fig:cmpscarce4}
}%
\caption{\textbf{Prediction performance (RMSE$_{\,i}$) on CMAPSS with simulated data scarcity \textit{in both train and test sets}.} The \textit{Posteriori Estimates} are outputs from ResGUR and AER, with the \textit{Predictions} rectified through the PR process.}
\label{fig:cmpscarce}
\end{figure*}

\subsection{Evaluation Metric}

\textbf{RMSE}
Root mean squared error (RMSE) is a commonly used metric for RUL prediction. 
For CMAPSS, where we only predict the RUL $y^{i'}_{T_{i'}}$ for each $i'$-th unit at $T_{i'}$, the RMSE can be calculated at subject level using the equation:
\begin{equation}
\text{RMSE}_{\,i}=(\frac{1}{N'}\sum_{i'=1}^{N'} (\hat y^{i'}_{T_{i'}} - y^{i'}_{T_{i'}})^2)^{\frac{1}{2}},
\end{equation}
where $N'$ is the total number of units in the testing data. For N-CMAPSS, we can calculated RMSE$_{\,t,s}$ and RMSE$_{\,t}$ given \textit{sample-wise} and \textit{interval-wise} predictions.
Another way to calculate the RMSE is denoted as $\text{RMSE}_{\,i,t}$,
which is calculated as the average RMSE$_{\,t}$ computed for each subject, accommodating variations in sample size across subjects. 

\textbf{$s\text{-score}$} 
Another commonly used metric in RUL prediction is the following scoring function:
\begin{equation}
s\text{-score}_{\,i}=
\begin{cases}
    \sum^{N'}_{i'=1} (\exp(-\frac{\Delta^{i'}_{T_{i'}}}{\tau_1}) - 1),& \Delta^{i'}_{T_{i'}}<0,\\
    \sum^{N'}_{i'=1} (\exp(\frac{\Delta^{i'}_{T_{i'}}}{\tau_2}) - 1), & \Delta^{i'}_{T_{i'}}\geq 0,
\end{cases}
\label{eq:s}
\end{equation}
where $\Delta^{i'}_{T_{i'}}=\hat y^{i'}_{T_{i'}} - y^{i'}_{T_{i'}}$, and $\tau_1=13$, $\tau_2=10$, to give greater penalty to the overestimated $\hat y^{i'}_{T_{i'}}$. 

\subsection{Settings}

\textbf{Model Configurations}
For FD002 and FD004, we use AER-$f$ with $\mathbf{W}_e$ sized $[R,32, 16, 8]$, $\mathbf{W}_d$ sized $[8,16,32,R]$, and $\mathbf{W}_o$ sized $[8,1]$, with \textit{learning rate}, $i.e.$, $\lambda$, of $10^{-2}$, and $B=100$ for FD002 and $B=200$ for FD004. $\text{ReLU}(\cdot)$ and a \textit{dropout rate} of $0.1$ are applied.
Both AER-$f$ and MLP-$f$ are used for N-CMAPSS, with $\lambda=10^{-2}$ and $B=1000$ for all subsets.
We also use ResGUR-$f$ with 4 layers of gated units, each of size $100$, followed by an MLP of size $[16,8,1]$, on FD001 and FD003, with $\lambda=10^{-3}$ and $B=100$.
Adam optimizer is adopted. 

\textbf{Parametric Labeling Function}
We use Equation \ref{eq:weibull} for labeling each sample $\mathbf{X}^i_t$ in CMAPSS, where $\beta=5$, $\theta_i=T_i/1.7$, and $\alpha=130$ following the common practice, 
which is applied to both $y^i_t$ and $\hat y^i_{t,s} \Leftrightarrow \hat y^i_{t}$. And linear RUL is already provided in N-CMAPSS as in Equation \ref{eq:linear} for both training and testing data, $i.e.$, $y^i_{t,s}=T_i-t$ and $y^{i'}_{t,s}=T_{i'}-t$.

\begin{table}[!t]
\caption[Prediction performance ($\text{RMSE}_{\,i,t}$) on mean N-CMAPSS with no scarcity simulated]{\textbf{Prediction performance ($\text{RMSE}_{\,i,t}$) on mean N-CMAPSS with no scarcity simulated}, compared to baseline work kNNI with LSS as in~\cite{viale2023least}.}
\centering\fontsize{9}{11}\selectfont
\begin{tabular}{>{\centering}p{1.2cm}>{\centering}p{1.8cm}>{\centering}p{1.5cm}>{\centering\arraybackslash}p{1.5cm}} 
\toprule
Method &  kNNI + LSS & $\textbf{MLP-$f$}$ &  $\textbf{AER-$f$}$ \\
\midrule
RMSE$_{\,i,t}$ & $7.43$ & $6.84_{\,\pm\,0.22}$ & $\textbf{6.49}_{\,\pm\,\textbf{0.04}}$\\
\bottomrule
\end{tabular}
\centering
\label{tab:dsmean}
\end{table}

\subsection{Experiment Results}

In all experiments, we repeated 10 times with different random seeds to evaluate uncertainty and conducted experiments with 2.6 GHz 6-Core Intel Core i7 CPU.

\subsubsection{Scarcity in Train Set Only (Q1)}
To compare with prior art~\cite{wang2021deep}, we evaluated prediction performance using 50\%, 30\%, and 10\% of samples from each subject in the train data of CMAPSS for model training, while reserving 100\% of the test data for prediction. Sampling was randomly conducted for each unit, resulting in a certain percentage of \textit{sparsely and irregularly} distributed samples along the timeline. The results are summarized in Table \ref{tab:cmscarce}. Our proposed methods achieve among the best performance in terms of RMSE$_{\,t}$, it shows even better performance or narrows the performance margin under more significant data scarcity.


\begin{figure}[!t]\centering
\subfigure[Impacts from Scarcity]{
\centering
\includegraphics[width=1.6in]{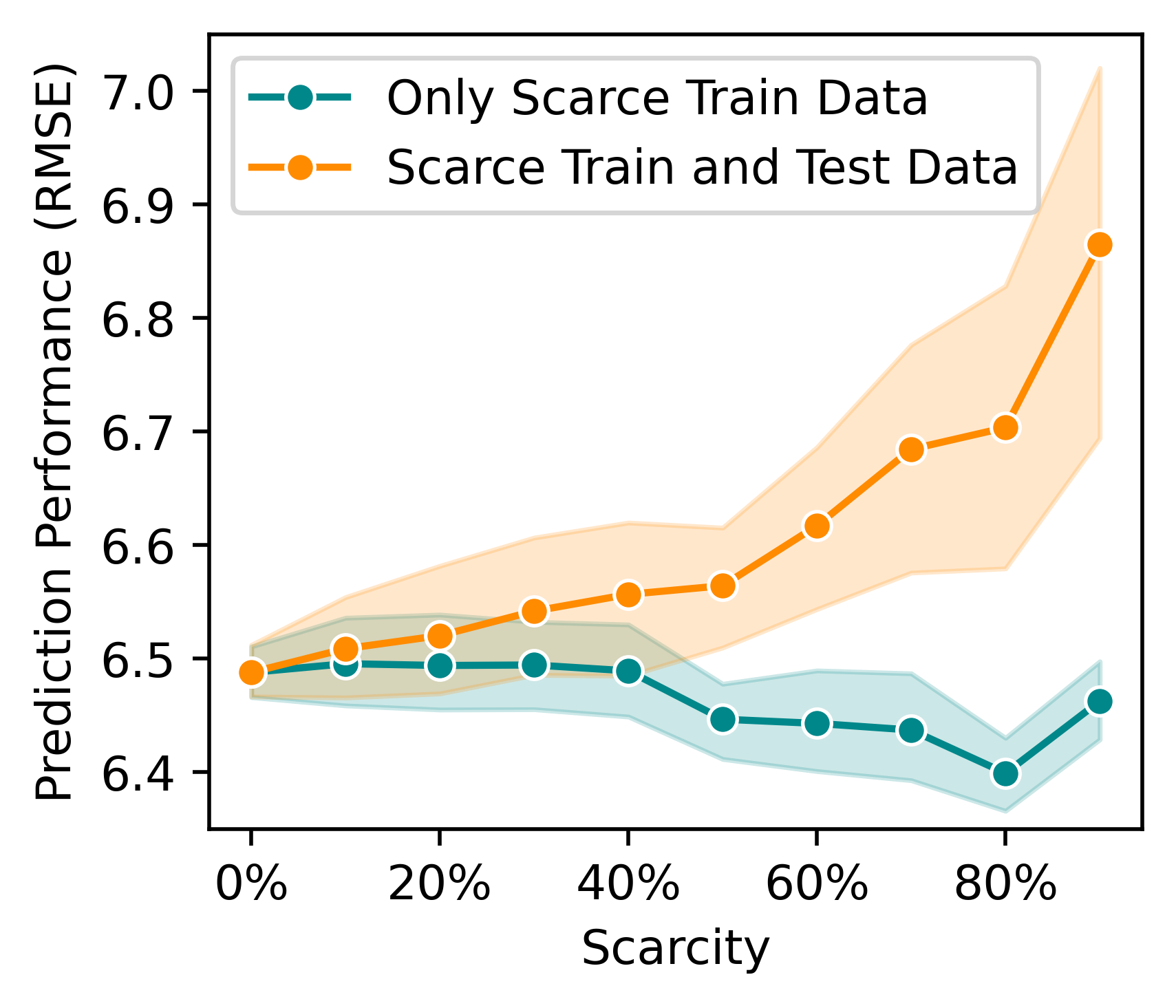}
\label{fig:meanncmap}
}%
\hfil
\subfigure[Unit 10 in DS08c]{
\centering
\includegraphics[width=1.4in]{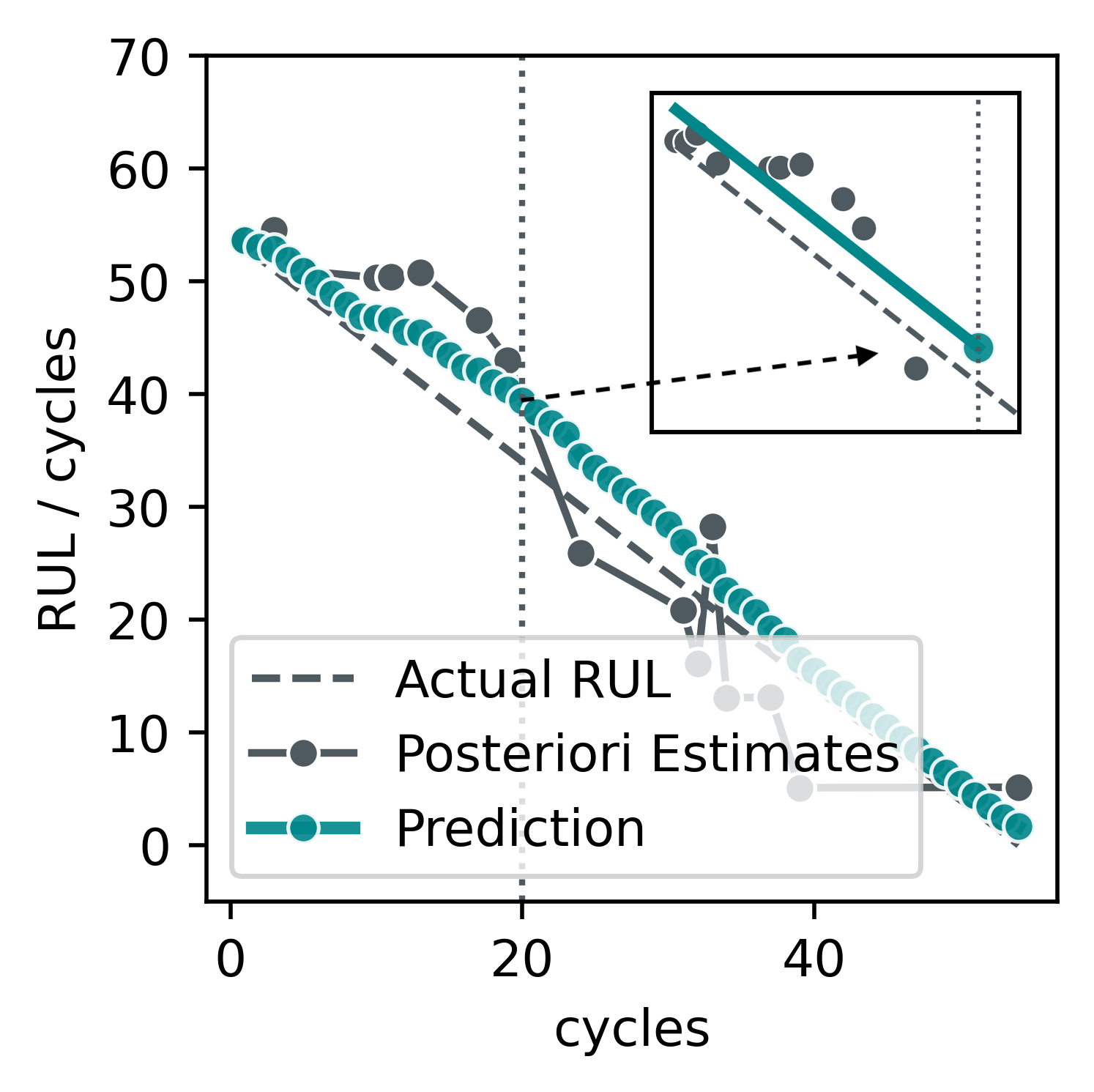}
\label{fig:8c-10}
}%
\caption{Prediction performance ($\text{RMSE}_{\,i,t}$) on mean N-CMAPSS with (a) simulated scarcity and (b) example demonstrations under 70\% scarcity similated.}
\label{fig:ncmapmeanscarce}
\vspace{-1em}
\end{figure}

\subsubsection{Scarcity in Both Train and Test Sets (Q1)}
However, in practical scenarios, data scarcity is often aligned in both training and testing data, a factor overlooked by prior studies. We conducted experiments using both train and test data with aligned data scarcity. The results on CMAPSS are presented in Figure \ref{fig:cmpscarce}.
AER-$f$ achieves more stable performance with increasing scarcity on FD002 and FD004. 

We also conducted data scarcity experiments with N-CMAPSS. 
By converting inputs as $\tilde{\mathbf{x}}^{i'}_{t}$ with mean values at $t$, the data volume is significantly reduced and we combine all subsets named as the \textbf{\textit{mean N-CMAPSS}} dataset. Prediction performance without scarcity is provided in Table \ref{tab:dsmean} and our method outperforms the previous work. 
The results with simulated scarcity are shown in Figure \ref{fig:meanncmap} under both scenarios of \textit{Only Scarce Train Data} and \textit{Scarce Train and Test Data}. Interestingly, the prediction performance improves when higher scarcity is simulated only in the train data, potentially due to noisy samples being removed. However, when scarcity is simulated in both train and test data, the performance degrades, while $\text{RMSE}_{\,i,t}$ remains below 7. Figure \ref{fig:8c-10} illustrates how our proposed method can predict current target values by leveraging historical posteriori estimates.

\subsection{Ablation Study}

\subsubsection{Effectiveness of PR Process (Q2)} 
Figure \ref{fig:cmpscarce} demonstrates the effectiveness of the PR process in enhancing the prediction performance of static regression models. While the \textit{Posteriori Estimates} are the direct outputs from the static models, the \textit{Predictions} represent the rectified results after applying the PR process. The results show that although the accuracy of the rectified predictions decreases with increasing data scarcity, the posteriori estimates remain largely unaffected. Notably, AER-$f$ exhibits significantly improved performance due to the PR process, even under 90\% data scarcity, as shown in Figures \ref{fig:cmpscarce2} and \ref{fig:cmpscarce4}.

\subsubsection{Impacts of Labeling Function (Q3)} 
Consistent with previous work, we employed a \textit{piece-wise linear function} for experiments with CMAPSS. The results in Table \ref{tab:label} demonstrate that using the proposed non-linear labeling function significantly enhances the performance.

\begin{table}[!t]
\caption{Prediction performance (RMSE$_{\,i}$) on FD002 \textbf{\textit{with varying labeling functions}} (AER$^\dagger$ using linear function) under simulated data scarcity \textit{in both train and test sets}. }
\centering\fontsize{9}{11}\selectfont
\begin{tabular}{>{\centering}p{1.4cm}>{\centering}p{1.7cm}>{\centering}p{1.7cm}>{\centering\arraybackslash}p{1.7cm}} 
\toprule
Scarcity &50\% & 70\% & 90\% \\
\midrule
AER$^\dagger$  &  $27.79_{\,\pm\,1.21}$ & $28.28_{\,\pm\,1.76}$ & $28.65_{\,\pm\,2.48}$ \\
AER-$f$  &  $\textbf{18.60}_{\,\pm\,1.60}$ & $\textbf{19.23}_{\,\pm\,1.91}$ & $\textbf{26.28}_{\,\pm\,2.48}$ \\
\bottomrule
\end{tabular}
\label{tab:label}
\end{table}

\begin{figure}[!t]\centering
\subfigure[Training Stage]{
\centering
\includegraphics[width=1.3in]{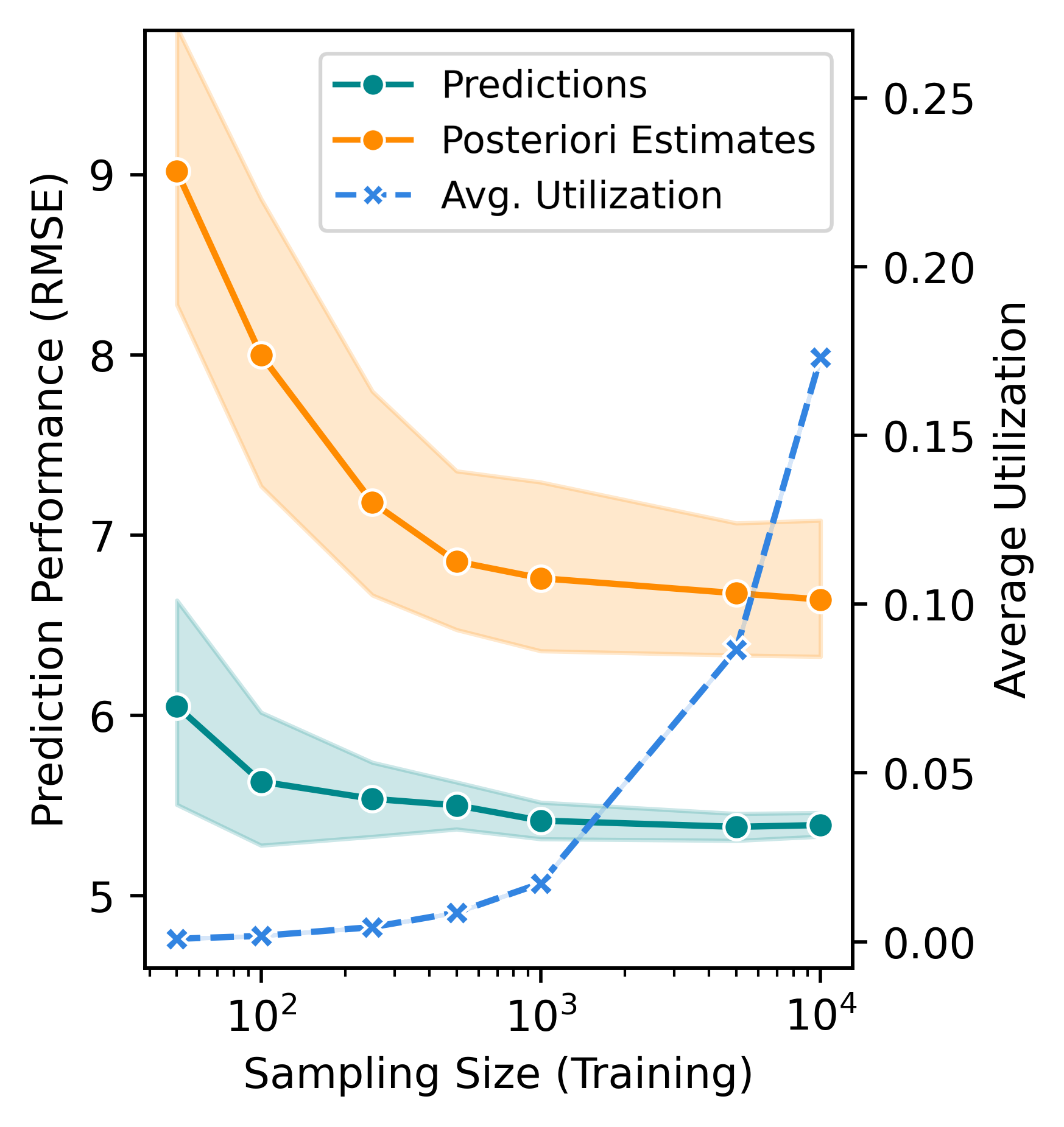}
\label{fig:sample_train}
}%
\hfil
\subfigure[Testing Stage]{
\centering
\includegraphics[width=1.35in]{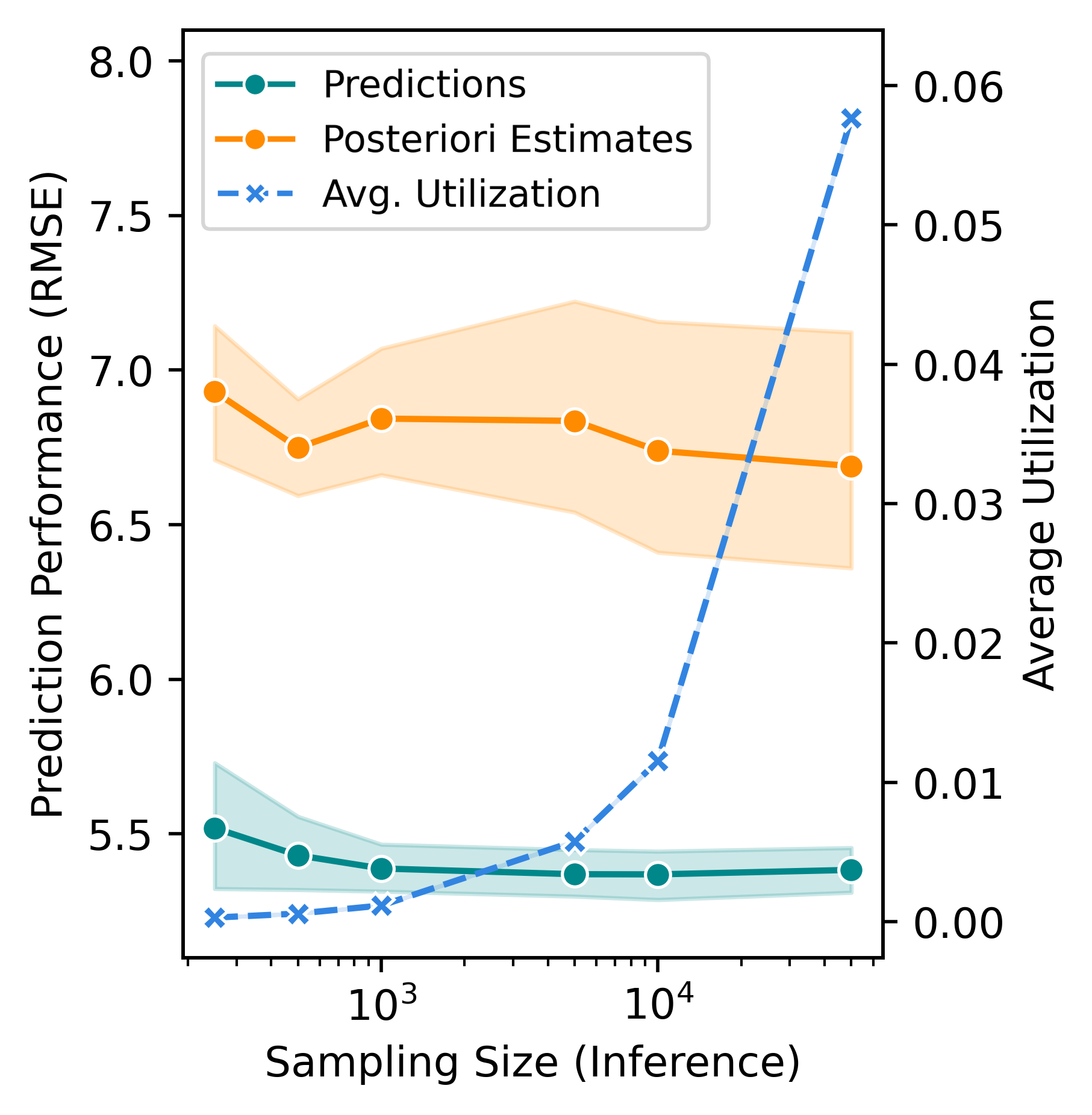}
\label{fig:sample_test}
}%
\caption{Prediction performance on DS02 \textbf{\textit{with varying sample size}}, $i.e.$, $B$, during (a) training and (b) testing stages.}
\label{fig:ablation}
\vspace{-1em}
\end{figure}

\subsubsection{Impacts of Reconstruction Loss in AER (Q4)} 
When $\gamma=0$, the AER model simplifies to a standard MLP model. As shown in Table \ref{tab:dsmean}, introducing the reconstruction loss allows AER-$f$ to outperform MLP-$f$.

\subsubsection{Sample Size for Identical Batch Training (Q4)} 
Figure \ref{fig:ablation} illustrates the impact of \textit{sample size} on the performance during the proposed identical batch training. The train data in DS02 (\textit{RMTS}) contains 5.3M samples from only 6 subjects. The figure demonstrates that sampling 1,000 data points per subject during training is sufficient to achieve optimal performance, comparable to using larger sample sizes. This corresponds to utilizing approximately 2\% of the total available data. Similar results are observed during testing.

\section{Conclusion}\label{sec:sum}

In this paper, we propose a unified framework, parameterized static regression, for RUL prediction with scarce time series data. Leveraging a parametric function for labeling target values, our models are trained efficiently and effectively with proposed batch training technique and achieves high prediction accuracy through PR process. Our framework uses data points instead of sequential inputs, inherently addressing data scarcity without special treatments. Temporal dependency is captured by leveraging historical estimates and approximating a parametric function based on underlying assumptions during labeling. Extensive experiments on RUL prediction with public benchmarks demonstrate the ability of proposed model to handle data scarcity with stable performance. 


%

%

\bibliographystyle{named}
\bibliography{mybibfile}

\end{document}